%% file: template.tex
\documentclass{article}
\usepackage{iclr2025_conference,times}
\input{math_commands.tex}

\usepackage{graphicx,booktabs,subcaption,amsmath,amssymb}
\usepackage{hyperref,url}
\hypersetup{hidelinks}
\iclrfinalcopy
\renewcommand{\eqref}[1]{(\ref{#1})}

\title{Evidence-Calibrated Selective Decoding for Reliable Visual Question Answering}
\author{Shiyu Gu, Luis Miguel Santana, Rafael Enrique Cruz\\
Technological University of the South\\
\texttt{rafael.cruz@utesur.edu.do}}
\begin{document}
\maketitle
\input{main}
\bibliography{references}
\bibliographystyle{iclr2025_conference}
\end{document}

%% file: math_commands.tex
\usepackage{amsmath,amsfonts,bm}

\def\eqref#1{equation~\ref{#1}}
\def\1{\bm{1}}

\DeclareMathAlphabet{\mathsfit}{\encodingdefault}{\sfdefault}{m}{sl}
\SetMathAlphabet{\mathsfit}{bold}{\encodingdefault}{\sfdefault}{bx}{n}

%% file: main.tex
\begin{abstract}
Visual question answering systems can produce confident responses that are weakly supported by their input images. Contrastive decoding can reduce this dependence on language priors, but a fixed correction may also overturn a correct answer, while selective answering can conceal errors by reducing coverage. We develop evidence-calibrated selective decoding, a framework that separates three decisions: whether local visual evidence supports a candidate, how strongly to correct the original distribution, and whether the resulting answer should be released under a verification budget. Evidence Verification across Crops (EVC) measures visual support and disagreement; Adaptive Contrast Correction (ACC) applies a bounded modification within the backbone's plausible candidate set; Budgeted Selective Answering (BSA) calibrates an answer-admission threshold on held-out images. We provide a complete mathematical specification, bounded-perturbation analysis, and an evaluation protocol that distinguishes forced answering from selective risk. Our quantitative analysis re-examines published baselines on POPE instantiated over MSCOCO, A-OKVQA, and GQA, together with four MME subtasks. These public measurements reveal variation across backbones, tasks, and correction settings that motivates adaptive control. The proposed framework has not yet undergone model-level evaluation; its effectiveness remains a falsifiable hypothesis rather than an established empirical result.
\end{abstract}

\section{Introduction}
A visual question answering system must distinguish what an image supports from what a language model finds plausible. A person standing near a dining table provides a strong linguistic cue for a plate or a cup, but neither object is guaranteed to be visible. A fluent affirmative response can therefore reflect a learned co-occurrence rather than a grounded observation. This distinction becomes especially consequential when the system must answer short questions, because a single unsupported word can determine the entire prediction. Reliable generation requires an explicit account of evidence, intervention, and the option to defer.

Visual in-context learning demonstrates that the information supplied to a large vision-language model can shape its behavior without retraining the whole model \citep{zhou2024visual}. Nevertheless, a useful context and a trustworthy decision are different requirements. Selecting influential contexts, as studied by GATEAU \citep{si2025gateau}, addresses which examples matter for alignment; at inference time, a related question is which parts of an image actually support the next decision. Our focus is the latter problem. We retain the frozen backbone and place a controlled verification mechanism around its next-token distribution.

\begin{figure}[!t]
\centering
\includegraphics[width=\linewidth]{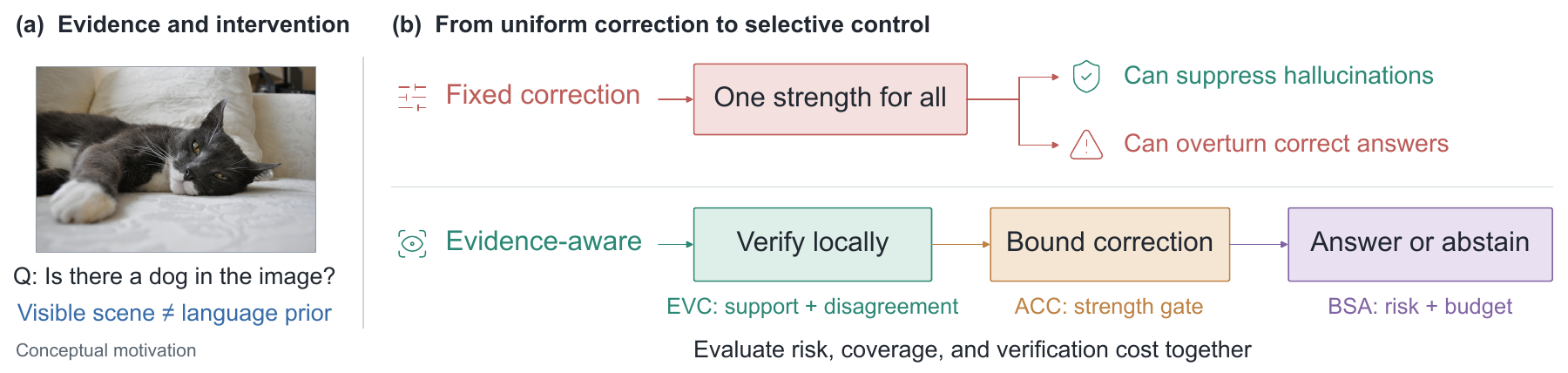}
\caption{Motivation: visual support, correction strength, and answer admission are separate decisions. Uniform intervention can suppress hallucinations while damaging correct answers. Conceptual illustration; no measured prediction trace. Photograph: Christoffer G. Thomsen, Wikimedia Commons, CC0.}
\label{fig:intro}
\end{figure}

Figure~\ref{fig:intro} contrasts the risk of a uniform intervention with the proposed separation of evidence verification, bounded correction, and selective release.

Visual contrastive decoding (VCD) compares predictions from an original image and a distorted counterpart \citep{vcd2024}. OPERA instead uses attention-related penalties and retrospection during decoding \citep{opera2024}. Both approaches motivate intervention after training, but neither the existence of a useful contrast nor a lower overall hallucination score establishes that every individual correction is beneficial. The intervention can suppress valid language structure, damage a correct answer, or fail when its comparison view removes relevant visual information. A strong verifier should be able to leave a prediction unchanged when additional evidence is uninformative.

Recent work reinforces this concern. Risk-aware selective prompting reports that verification can help difficult examples while introducing errors on easier ones \citep{rsp2026}. Claim admission moves the decision from token correction toward evidence-dependent acceptance \citep{claim2026}. These developments make a broad claim of novelty for visual comparison or abstention untenable. The question pursued here is narrower: can local evidence disagreement, bounded token correction, and budget-aware answer admission be specified as a coherent and testable inference procedure?

We propose three coupled components. EVC scores candidate tokens using multiple image views and a visual-null reference. ACC scales a clipped correction by the agreement of these views and the original model's dependence on its visual input. BSA releases an answer only when its risk score passes a threshold selected on independent calibration images. The verification budget bounds extra model evaluations, and all evaluations must report coverage alongside error. This design makes it possible to distinguish a useful correction from an apparent improvement caused solely by answering fewer questions.

The paper makes three contributions. First, it gives an operational definition of local visual evidence that distinguishes missing evidence from evidence of absence. Second, it derives a bounded correction with an exact fallback to the backbone and states the assumptions needed for finite-grid calibration. Third, it audits published baseline values and specifies a matched-checkpoint evaluation, including component ablations, parameter sweeps, distribution shifts, and failure cases. The numerical results in this manuscript belong to the cited baseline publication; we do not attribute them to our proposed method. No model checkpoint was executed for this study, so the manuscript should be read as a method and evaluation study with an explicit empirical boundary.

\section{Related Work}
\subsection{Visual conditioning, feedback, and generalization}
Large vision-language models combine visual representations with the generative capabilities of language models. BLIP-2 uses a lightweight bridge between pretrained components \citep{blip2023}, while visual instruction tuning adapts the interface to conversational tasks \citep{llava2023}. InstructBLIP broadens instruction-conditioned visual understanding \citep{instruct2023}; MiniGPT-4 \citep{minigpt2023} and mPLUG-Owl \citep{mplug2024} provide additional examples of multimodal alignment. These systems differ in training data, architecture, and decoding behavior, so their published scores cannot serve as a controlled ablation of an inference algorithm.

Weak-to-strong generalization across multiple capabilities \citep{zhou2025weak} motivates separating the quality of supervision from the capability of the eventual model. Our framework applies a related caution to inference: a weak verifier can be useful only if its limitations are represented rather than hidden behind a hard decision. Abnormal-aware feedback for medical vision-language models \citep{zhou2025improving} similarly highlights the importance of task-specific error information. We cite that work as a feedback-related connection, without claiming that the present natural-image protocol validates clinical behavior or transfers directly to medical images.

Entropy-guided optimization for autoregressive image generation \citep{songbroad} studies control over a generative process. Here uncertainty is used at inference to regulate evidence acquisition and correction; it is not a replacement for that training objective. These connections explain the relation to our source literature while preserving the task shift from representation learning and visual generation to answer reliability.

\subsection{Faithfulness and contrastive inference}
Faithfulness-aware decoding in summarization \citep{wan-etal-2023-faithfulness} and contrastive decoding in multilingual translation \citep{waldendorf-etal-2024-contrastive} show that decoding choices can influence consistency with a conditioning source. Their findings motivate source-aware decisions, but textual consistency is not equivalent to visual correctness. A crop can alter the semantic content of the source itself. Contrastive explanations \citep{yin-neubig-2022-interpreting} also motivate examining changes between conditions, although such changes alone do not identify a causal visual feature.

VCD supplies a direct baseline for visual logit contrast \citep{vcd2024}; OPERA offers an alternative based on over-trust and retrospection \citep{opera2024}. Context Embedding Injection examines layer-wise generation and injects grounding information from the input context \citep{cei2026}. Causal Decoding targets spurious dependencies during generation \citep{causal2026}. SAFE uses visual ablation as a probe of grounding failures \citep{safe2026}. These methods act at different points in the inference process. We distinguish our proposal through the explicit combination of view disagreement, a bounded correction, and calibrated selective release, rather than asserting that its ingredients are individually new.

Training for contextual faithfulness with synthetic tasks and reinforcement learning \citep{DBLP:conf/aaai/SiZGBWGLLHCQZCS26} addresses model behavior through a different intervention point. Our frozen-backbone setting is complementary: it asks how much can be achieved without changing parameters, while explicitly accounting for the additional inference cost. It cannot replace a training method when the backbone lacks the visual representation required to answer a question.

\subsection{Hallucination evaluation and selective reporting}
POPE measures object hallucination through balanced binary queries and several negative-object sampling schemes \citep{li-etal-2023-evaluating}. CHAIR evaluates unsupported object mentions in image captions \citep{chair2018}, and HaluEval studies hallucination evaluation for language models \citep{li-etal-2023-halueval}. These metrics cover different output spaces; a gain on one should not be advertised as universal faithfulness. MME evaluates both perception and cognition, with task scores based on answer accuracy and paired-question consistency \citep{mme2025}. CDH-Bench adds a recent diagnostic of conflict between visual evidence and commonsense \citep{cdh2026}.

Selective release further changes the estimand. A method that answers half the questions can reduce error among released answers without improving forced-answer accuracy. Risk-aware prompting and claim admission make the importance of this distinction concrete \citep{rsp2026,claim2026}. We therefore require separate forced-answer scores, coverage-conditioned comparisons, and rejection accounting. The finite-grid certificate below controls risk only under its stated sampling and independence assumptions; it does not establish robustness to arbitrary dataset shift.

\section{Method}
\subsection{Task, notation, and design boundary}
Let $x$ be an image, $q$ a question, and $h_t=(q,y_{<t})$ the textual history at decoding step $t$. The frozen backbone produces a distribution over vocabulary $\mathcal V$:
\begin{align}
p_t^0(w) = p_\theta(w\mid x,h_t), \qquad w\in\mathcal V .
\label{eq:base}
\end{align}
Equation~\eqref{eq:base} defines the distribution to which the procedure must return when verification is disabled. We use an empty visual embedding, processed by the same implementation, to obtain $p_t^{\varnothing}$. The exact null construction must be logged because a black image, a zero embedding, and a missing-image prompt can produce different distributions.

We retain only candidates plausible under the original model:
\begin{align}
\mathcal H_t=\{w:p_t^0(w)\geq\beta\max_{u\in\mathcal V}p_t^0(u)\},\quad 0<\beta\leq1 .
\label{eq:head}
\end{align}
The constraint in Eq.~\eqref{eq:head} limits the correction's ability to promote an originally implausible token, as in plausibility-constrained contrastive decoding. It also limits recall: if the correct token is outside this set, the procedure cannot recover it. This is a deliberate, testable restriction rather than a guarantee of correctness.

\begin{figure}[!t]
\centering
\includegraphics[width=\linewidth]{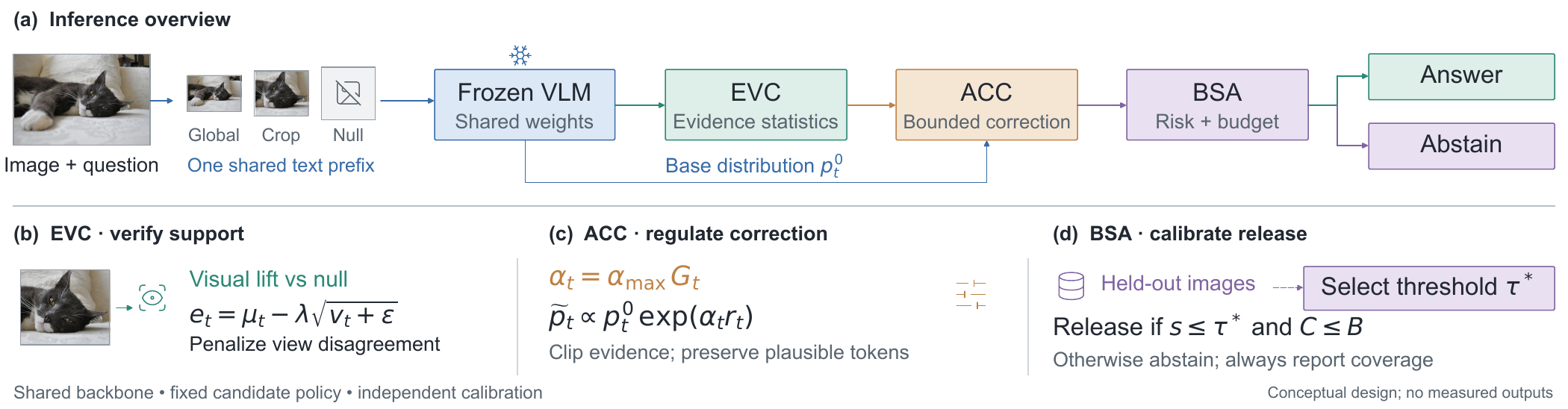}
\caption{Inference overview and mechanism details. EVC verifies support, ACC bounds correction, and BSA controls release. Green rectangle: matched crop; dashed arrow: calibration. All views share one frozen backbone and prefix. Conceptual design; photo and crops: Christoffer G. Thomsen, CC0.}
\label{fig:architecture}
\end{figure}

Figure~\ref{fig:architecture} summarizes the proposed pipeline. We refer to the combined system as evidence-calibrated selective decoding and to its full configuration as EVC+ACC+BSA. The visual branches and the language model remain frozen. The approach is training-free with respect to backbone weights, but its answer-admission threshold uses labeled calibration data; that requirement must be stated when comparing it with entirely uncalibrated baselines.

\subsection{EVC: evidence verification across crops}
A fixed view generator retains the global image and constructs local crops:
\begin{align}
\mathcal X(x)=\{x^{(0)}=x,\ x^{(k)}=\operatorname{crop}(x,b_k)\}_{k=1}^{K} .
\label{eq:views}
\end{align}
The boxes in Eq.~\eqref{eq:views} are determined by a grid before test labels are inspected. An attention-based selector is an optional ablation, not the default. Keeping a global view is necessary for relational questions, because a crop can remove a referent while retaining the object being queried. Each crop uses the backbone's native image preprocessing; independently reinterpreting the same question for each crop is prohibited.

For a candidate $w$, EVC computes a clipped visual lift relative to the null condition:
\begin{align}
d_{t,k}(w)=\operatorname{clip}\!\left(\log\frac{p_\theta(w\mid x^{(k)},h_t)+\epsilon}{p_t^{\varnothing}(w)+\epsilon},-c,c\right) .
\label{eq:lift}
\end{align}
Equation~\eqref{eq:lift} is an operational support statistic, not a causal effect. Null inputs can be out of distribution, and a token may be supported by contextual regularities shared across all views. The small fixed $\epsilon$ prevents numerical singularities; sensitivity to the clipping bound $c$ must be reported.

We aggregate support conservatively:
\begin{align}
\mu_t(w)&=\frac{1}{K+1}\sum_{k=0}^{K}d_{t,k}(w),\quad
v_t(w)=\frac{1}{K+1}\sum_{k=0}^{K}(d_{t,k}(w)-\mu_t(w))^2,\nonumber\\
e_t(w)&=\mu_t(w)-\lambda\sqrt{v_t(w)+\epsilon} .
\label{eq:evidence}
\end{align}
The second line of Eq.~\eqref{eq:evidence} penalizes disagreement. It is not a statistical confidence interval: the crops overlap and are strongly dependent. In particular, many background views may lack a small but real object. EVC must therefore be evaluated against the maximum-view and global-only alternatives rather than assuming that mean-minus-dispersion is always preferable.

To summarize disagreement at the distribution level, we restrict each view distribution to $\mathcal H_t$, normalize it as $\bar p_{t,k}$, and form their mean $\bar p_t$:
\begin{align}
D_t=\frac{1}{K+1}\sum_{k=0}^{K}\operatorname{KL}(\bar p_{t,k}\,\|\,\bar p_t),\qquad
\bar p_t=\frac{1}{K+1}\sum_{k=0}^{K}\bar p_{t,k} .
\label{eq:disagreement}
\end{align}
Unlike the scalar support of an individual candidate, Eq.~\eqref{eq:disagreement} captures variation across the entire retained distribution. Low disagreement is insufficient evidence by itself: the model can make the same prior-driven mistake across every crop. We consequently keep evidence strength and disagreement as separate quantities.

\subsection{ACC: bounded adaptive contrast correction}
We first measure whether the image changes the original distribution relative to its null counterpart, with both normalized over the same candidate set:
\begin{align}
J_t=\operatorname{JS}(\bar p_t^0,\bar p_t^{\varnothing}),\qquad
G_t=\sigma\!\left(a(J_t-j_0)\right)\exp(-bD_t) .
\label{eq:gate}
\end{align}
The gate in Eq.~\eqref{eq:gate} becomes small when the image barely affects the distribution or when crop predictions disagree. This is a conservative choice: rather than increasing correction precisely when evidence is most uncertain, it asks BSA to consider abstention. The alternative policy of allocating additional verification to high-disagreement examples belongs in the budget ablation.

The correction strength and centered evidence are bounded:
\begin{align}
\alpha_t=\alpha_{\max}G_t,\qquad
r_t(w)=\operatorname{clip}(e_t(w)-e_t(w_t^0),-c,c),\quad
w_t^0=\arg\max_w p_t^0(w) .
\label{eq:strength}
\end{align}
Centering in Eq.~\eqref{eq:strength} makes the comparison interpretable relative to the original preferred token. It does not itself change a softmax if clipping is inactive, but it does affect which differences reach the clipping boundary. The null and crop computations use the same prefix; otherwise a score difference could reflect changed textual history rather than changed visual conditioning.

ACC constructs the modified distribution as
\begin{align}
\widetilde p_t(w)=
\frac{\mathbf1\{w\in\mathcal H_t\}\exp(\log p_t^0(w)+\alpha_t r_t(w))}
{\sum_{u\in\mathcal H_t}\exp(\log p_t^0(u)+\alpha_t r_t(u))} .
\label{eq:corrected}
\end{align}
Equation~\eqref{eq:corrected} separates candidate restriction from evidence correction. For any two retained tokens, the bounded perturbation gives
\begin{align}
\left|\log\frac{\widetilde p_t(u)}{\widetilde p_t(v)}-
\log\frac{p_t^0(u)}{p_t^0(v)}\right|\leq2\alpha_{\max}c .
\label{eq:bound}
\end{align}
The proof of Eq.~\eqref{eq:bound} follows by canceling the shared normalizer and bounding $\alpha_t|r_t(u)-r_t(v)|$. It limits how much an intervention can alter relative odds; it does not imply a reduction in error. If the backbone is wrong by a large margin, this protection can prevent a necessary change.

Disabling correction must also disable candidate truncation when reproducing the exact backbone:
\begin{align}
\alpha_{\max}=0,\quad\mathcal H_t=\mathcal V
\quad\Longrightarrow\quad\widetilde p_t=p_t^0 .
\label{eq:fallback}
\end{align}
Equation~\eqref{eq:fallback} is an implementation invariant. Setting only $\alpha_{\max}=0$ while leaving a truncated set would yield a conditional version of the backbone distribution, not exact equality. This detail matters when reporting a zero-strength baseline.

Local optimization with representation continuity in autoregressive video generation \citep{zhou2026accelerating} provides a related motivation for avoiding unnecessary disruption across sequential steps. We use the connection as a design analogy: the present odds bound concerns inference distributions, not a claim that its theoretical behavior follows from a video-training result.

\subsection{BSA: budgeted selective answering}
For a completed candidate answer $y$, let $\mathcal T(y)$ denote the steps selected for verification under a fixed token-selection rule. With binary POPE questions, the rule evaluates the explicit yes/no answer margin; it does not infer object absence from a missing crop. For longer answers it can aggregate designated content-token positions, which must be logged. Define a bounded risk score
\begin{align}
s(x,q,y)=\sigma\!\left(\eta_0+\eta_1\max_{t\in\mathcal T(y)}D_t
-\eta_2\min_{t\in\mathcal T(y)}e_t(y_t)
-\eta_3 M(y)\right) ,
\label{eq:risk-score}
\end{align}
where $M(y)$ is a normalized answer margin. The nonnegative coefficients in Eq.~\eqref{eq:risk-score} are fixed before the final calibration partition is used. The score need not be a calibrated probability; the next stage assesses the error of the admitted set directly. A broad planner objective, as studied in long-horizon agent training \citep{si2026goal}, is another reason to keep action scheduling distinct from local token scoring, although our task is limited to image-conditioned answers.

Let $C(x,q)$ be the total verification cost in extra forward equivalents, determined independently of the chosen admission threshold. The policy is
\begin{align}
A_\tau(x,q)=\mathbf1\{s(x,q,y)\leq\tau\}\,
\mathbf1\{C(x,q)\leq B\} .
\label{eq:admission}
\end{align}
The answer is released only when Eq.~\eqref{eq:admission} holds. Exceeding the budget produces an explicit abstention, not a negative answer. A fixed schedule determines verification views so that changing $\tau$ does not change which evidence was collected; this keeps calibration policies fixed.

On an independent calibration set, write $n_\tau$ for the number admitted and $E_i$ for a binary answer error. The empirical selective risk is
\begin{align}
\widehat R_\tau=\frac{\sum_{i=1}^{n}A_\tau(x_i,q_i)E_i}{n_\tau},\qquad
n_\tau=\sum_{i=1}^{n}A_\tau(x_i,q_i) .
\label{eq:empirical}
\end{align}
Equation~\eqref{eq:empirical} is defined only for nonempty admitted sets. Repeated questions from one image are dependent, so the certificate uses one prespecified calibration query per independently sampled image. Other questions may be used for descriptive evaluation, but not counted as independent observations in this bound.

For a threshold grid of size $L$ fixed independently of calibration labels, a Hoeffding union bound gives
\begin{align}
U_\tau=\min\!\left\{1,\widehat R_\tau+
\sqrt{\frac{\log(L/\delta)}{2n_\tau}}\right\},\qquad n_\tau>0 .
\label{eq:ucb}
\end{align}
For empty sets we define $U_\tau=1$. Equation~\eqref{eq:ucb} is valid under independent sampling of the calibration units, a fixed error rule, and policies fixed before calibration labels are seen. Conditional on the admission count for a fixed policy, admitted errors are independent draws from that policy's accepted population. The union bound permits selection among the prespecified policies; it does not cover fitting the risk scorer on the same labels.

We choose the highest-coverage feasible policy:
\begin{align}
\tau^*=\arg\max_{\tau\in\mathcal G:\,U_\tau\leq r_*}\frac{n_\tau}{n} .
\label{eq:selection}
\end{align}
Ties in Eq.~\eqref{eq:selection} use the smaller threshold. If there is no feasible threshold, BSA abstains on every query and reports zero coverage. Under the stated assumptions, the risk of the chosen nonempty policy is at most $r_*$ with probability at least $1-\delta$. This statement concerns the calibration-generating distribution; changing the image domain, model checkpoint, or decoding rule invalidates its direct application.

\subsection{Cost, implementation, and limits of the design}
Separating evidence acquisition from release gives a simple compute accounting. If $T$ steps use the base path and $T_v$ verified steps each evaluate $K$ crops and one null path, a conservative accounting is
\begin{align}
C_{\mathrm{total}}\leq T C_{\mathrm{base}}+
T_v(K+1)C_{\mathrm{aux}}+C_{\mathrm{score}} .
\label{eq:cost}
\end{align}
Equation~\eqref{eq:cost} counts complete auxiliary calls; image-feature caching can reduce their constant cost, but their text histories still require compatible state updates. Simultaneously caching all view-specific key/value states may multiply memory usage. Sequential view evaluation reduces peak memory but can increase latency. Neither option is inherently free.

An implementation should first run the unmodified checkpoint and preserve its tokenization, prompt, and stopping behavior. It should then construct view-specific image features, teacher-force the common text prefix into each path, compute evidence on the same candidate IDs, and apply the clipped correction. The final release stage reads logged scores and the frozen calibration threshold. A request-level budget prevents repeated verification from silently accumulating across a long answer. Parameter sharing with the base model means no new backbone weights, but it does not mean zero additional computation or zero calibration data.

The design rests on hypotheses rather than universal properties. Crops may be useful for object presence yet destructive for counting or relative position. The null path may exaggerate contrast for out-of-distribution inputs. A low gate may preserve a correct answer or preserve a mistake. These alternatives motivate the ablations and failure analysis below and constrain what a successful empirical result would justify.

\section{Experimental Setup and Evidence Provenance}
\subsection{Datasets and task definitions}
We use four evaluation sources specified in Table~\ref{tab:setup}. MSCOCO provides common-object scenes \citep{coco2014}; A-OKVQA introduces knowledge-rich visual questions \citep{aokvqa2022}; GQA emphasizes compositional scene understanding \citep{gqa2019}. In the POPE instantiations used here, these image collections support binary existence questions, not their original open-ended VQA labels. Random, popular, and adversarial negative sampling are separate conditions. A score on a POPE instantiation must therefore be named with both its image source and probing protocol.

\begin{table}[!t]
\centering
\caption{Evaluation specification. The proposed-method settings are a preregisterable protocol, not a description of completed model experiments.}
\label{tab:setup}
\resizebox{\linewidth}{!}{\begin{tabular}{llll}
\toprule
Source & Evaluation task & Primary report & Separation rule\\
\midrule
MSCOCO + POPE & Binary existence, three samplers & F1, accuracy, Yes ratio & Split by image ID\\
A-OKVQA + POPE & Binary existence, three samplers & F1, accuracy, Yes ratio & Preserve original question set\\
GQA + POPE & Binary existence, three samplers & F1, accuracy, Yes ratio & Preserve original question set\\
MME & Existence, count, position, color & Native task scores, sum /800 & Keep paired questions together\\
All selective tests & Answer or abstain & Risk, coverage, cost & Calibration disjoint from test\\
\bottomrule
\end{tabular}}
\end{table}

The MME analysis uses only four hallucination-related subtasks. Their total has maximum 800, because each subtask contributes up to 200 from accuracy plus paired-question accuracy. This is different from the full perception score or the overall benchmark. The proposed risk certificate uses binary per-query correctness rather than the paired-question MME statistic. Both must be reported without treating one as a calibrated version of the other.

The POPE paper provides a useful original reference: its MSCOCO evaluation uses 500 images and six questions per image, balancing positive and negative labels \citep{li-etal-2023-evaluating}. The VCD paper supplies a consistent set of published means and standard deviations across three backbones and all three source collections \citep{vcd2024}. We use those measurements as historical anchors. They were not rerun locally, and they cannot establish the performance of any proposed component.

\subsection{Matched-checkpoint comparison}
The primary comparison should hold constant a frozen checkpoint, native image resolution, prompt, numerical precision, maximum answer length, test IDs, and answer parser. The baseline set comprises regular decoding, VCD, OPERA, CEI, SAFE, and RSP, together with EVC+ACC+BSA. Each intervention should be ported only where its required model internals are available. An incompatible method is documented separately; its score from another backbone is not inserted into a matched ranking.

VCD's official implementation provides the main integration point, and OPERA's official repository provides an additional decoding reference. Both paper--repository pairs were checked. Their code was not installed or benchmarked on a model in this workspace. A complete empirical release must record a commit hash, model revision, image preprocessing configuration, package versions, accelerator, and command line. A repository link by itself is insufficient evidence that a result is reproducible.

A development partition selects the view schedule, gate parameters, score coefficients, and candidate rule. A separate calibration partition selects only $\tau$ on a fixed grid; final test data choose none of these. Domains are also evaluated under transfer: calibrate on a COCO development source, then test the same frozen policy on A-OKVQA and GQA. The transferred certificate is reported as inapplicable unless its distributional assumptions are justified. A separate target-calibrated evaluation diagnoses how much of the failure is calibration shift.

\subsection{Metrics and statistical units}
Forced-answer evaluation reports accuracy, precision, recall, F1, and affirmative-response frequency for POPE. Abstention is not parsed as ``No.'' A separate forced-answer run disables BSA while retaining any token correction, making its effect identifiable. Selective evaluation reports empirical risk and coverage:
\begin{align}
\widehat{\operatorname{Cov}}=\frac1N\sum_{i=1}^{N}A_i,\qquad
\widehat{\operatorname{Risk}}=\frac{\sum_i A_iE_i}{\sum_i A_i} .
\label{eq:metrics}
\end{align}
Equation~\eqref{eq:metrics} makes the denominator change explicit. For an empty accepted set, risk is undefined and coverage is zero; the table should print the policy outcome rather than a misleading zero error. Fixed-coverage comparisons use 50\%, 70\%, 90\%, and full coverage where attainable. Area under the risk--coverage curve supplements these operating points but does not replace them.

Confidence intervals should resample images as clusters so that related questions remain together. Repeated decoding seeds and bootstrap uncertainty measure different variation and require separate labels. The original publication's standard deviations are reproduced as published; they are not confidence intervals estimated by us. A significance claim must use a paired comparison on the same test images, and method selection must not be based on the final paired test.

\section{Published Baseline Analysis}
\subsection{POPE across three image sources}
Tables~\ref{tab:coco}, \ref{tab:aok}, and \ref{tab:gqa} reproduce F1 values for six published model--decoding configurations per dataset from VCD Table~1. They are grouped by backbone and represent two decoding methods, not six independent algorithms. The reference includes sampling-based regular decoding; calling it a greedy baseline would change the setting. We retain the publication's means and standard deviations, including small variations that may look surprising, rather than smoothing the table into a preferred narrative.

For MSCOCO, the published LLaVA1.5 F1 improves from 81.33 to 87.16 under random negatives, a gain of 5.83 percentage points. Under adversarial negatives, the gain is 3.76 points. Qwen-VL gains 5.14 points for random negatives and 3.53 for adversarial negatives. InstructBLIP's corresponding gains are 3.27 and 2.93 points. The intervention helps all these published conditions, but its magnitude is not constant across the sampling strategy or backbone. These observations support testing an adaptive rule; they do not establish which adaptation should work.

\begin{table}[!t]
\centering
\caption{Published POPE F1 (\%) on MSCOCO, reproduced from VCD Table~1 \citep{vcd2024}. Standard deviations are those of the source publication. Rows are six model--decoding configurations, grouped by backbone; they do not report our method.}
\label{tab:coco}
\resizebox{\linewidth}{!}{\begin{tabular}{lllll}
\toprule
Backbone & Decoding & Random & Popular & Adversarial\\
\midrule
LLaVA1.5 \citep{llava2023} & Regular & $81.33\pm0.41$ & $80.06\pm0.05$ & $77.57\pm0.57$\\
LLaVA1.5 \citep{llava2023} & VCD & $87.16\pm0.41$ & $85.06\pm0.37$ & $81.33\pm0.34$\\
Qwen-VL \citep{vcd2024} & Regular & $82.67\pm0.41$ & $82.06\pm0.23$ & $80.37\pm0.37$\\
Qwen-VL \citep{vcd2024} & VCD & $87.81\pm0.11$ & $86.40\pm0.09$ & $83.90\pm0.39$\\
InstructBLIP \citep{instruct2023} & Regular & $80.41\pm0.80$ & $78.36\pm0.76$ & $76.59\pm0.40$\\
InstructBLIP \citep{instruct2023} & VCD & $83.68\pm0.40$ & $81.07\pm0.39$ & $79.52\pm0.38$\\
\bottomrule
\end{tabular}}
\end{table}

\begin{table}[!t]
\centering
\caption{Published POPE F1 (\%) on A-OKVQA, reproduced from VCD Table~1 \citep{vcd2024}. Standard deviations are those of the source publication. Rows are six model--decoding configurations, grouped by backbone; they do not report our method.}
\label{tab:aok}
\resizebox{\linewidth}{!}{\begin{tabular}{lllll}
\toprule
Backbone & Decoding & Random & Popular & Adversarial\\
\midrule
LLaVA1.5 \citep{llava2023} & Regular & $82.56\pm0.50$ & $79.59\pm0.37$ & $75.15\pm0.23$\\
LLaVA1.5 \citep{llava2023} & VCD & $86.34\pm0.21$ & $82.82\pm0.36$ & $77.73\pm0.29$\\
Qwen-VL \citep{vcd2024} & Regular & $85.59\pm0.53$ & $84.63\pm0.42$ & $79.50\pm0.38$\\
Qwen-VL \citep{vcd2024} & VCD & $89.01\pm0.16$ & $87.81\pm0.31$ & $82.38\pm0.10$\\
InstructBLIP \citep{instruct2023} & Regular & $81.86\pm0.32$ & $78.17\pm0.73$ & $75.56\pm0.57$\\
InstructBLIP \citep{instruct2023} & VCD & $84.56\pm0.28$ & $81.15\pm0.42$ & $77.19\pm0.47$\\
\bottomrule
\end{tabular}}
\end{table}

\begin{table}[!t]
\centering
\caption{Published POPE F1 (\%) on GQA, reproduced from VCD Table~1 \citep{vcd2024}. Standard deviations are those of the source publication. Rows are six model--decoding configurations, grouped by backbone; they do not report our method.}
\label{tab:gqa}
\resizebox{\linewidth}{!}{\begin{tabular}{lllll}
\toprule
Backbone & Decoding & Random & Popular & Adversarial\\
\midrule
LLaVA1.5 \citep{llava2023} & Regular & $82.95\pm0.28$ & $78.37\pm0.18$ & $76.06\pm0.24$\\
LLaVA1.5 \citep{llava2023} & VCD & $86.99\pm0.41$ & $82.24\pm0.35$ & $78.78\pm0.36$\\
Qwen-VL \citep{vcd2024} & Regular & $79.01\pm0.40$ & $74.84\pm0.34$ & $74.33\pm0.71$\\
Qwen-VL \citep{vcd2024} & VCD & $85.33\pm0.38$ & $82.23\pm0.22$ & $80.75\pm0.27$\\
InstructBLIP \citep{instruct2023} & Regular & $80.56\pm0.18$ & $76.42\pm0.52$ & $74.12\pm0.58$\\
InstructBLIP \citep{instruct2023} & VCD & $84.16\pm0.01$ & $80.17\pm0.16$ & $77.53\pm0.08$\\
\bottomrule
\end{tabular}}
\end{table}

A-OKVQA reveals a different balance. LLaVA1.5 improves by 3.78 points under random negatives but 2.58 under adversarial negatives. InstructBLIP's adversarial improvement is 1.63 points. This smaller increase should not be hidden by an average across sources. The questions still probe object presence; the values do not measure the models' original A-OKVQA knowledge-based answer generation. An inference procedure that uses scene context must be examined for both genuine visual support and inherited object co-occurrence.

\begin{figure}[!t]
\centering
\begin{subfigure}{0.48\linewidth}\centering
\includegraphics[width=\linewidth]{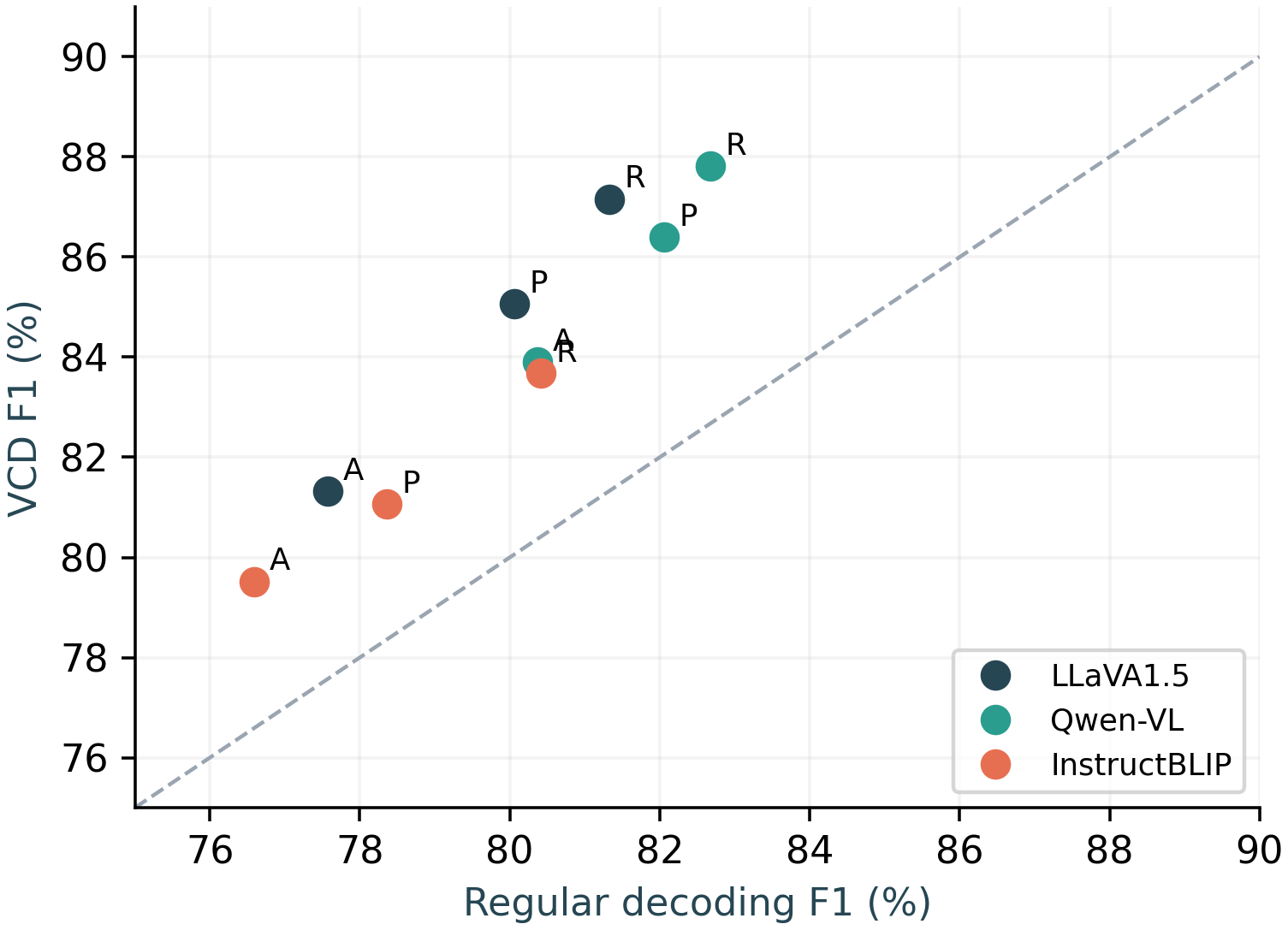}
\caption{MSCOCO F1 before and after VCD.}
\end{subfigure}\hfill
\begin{subfigure}{0.48\linewidth}\centering
\includegraphics[width=\linewidth]{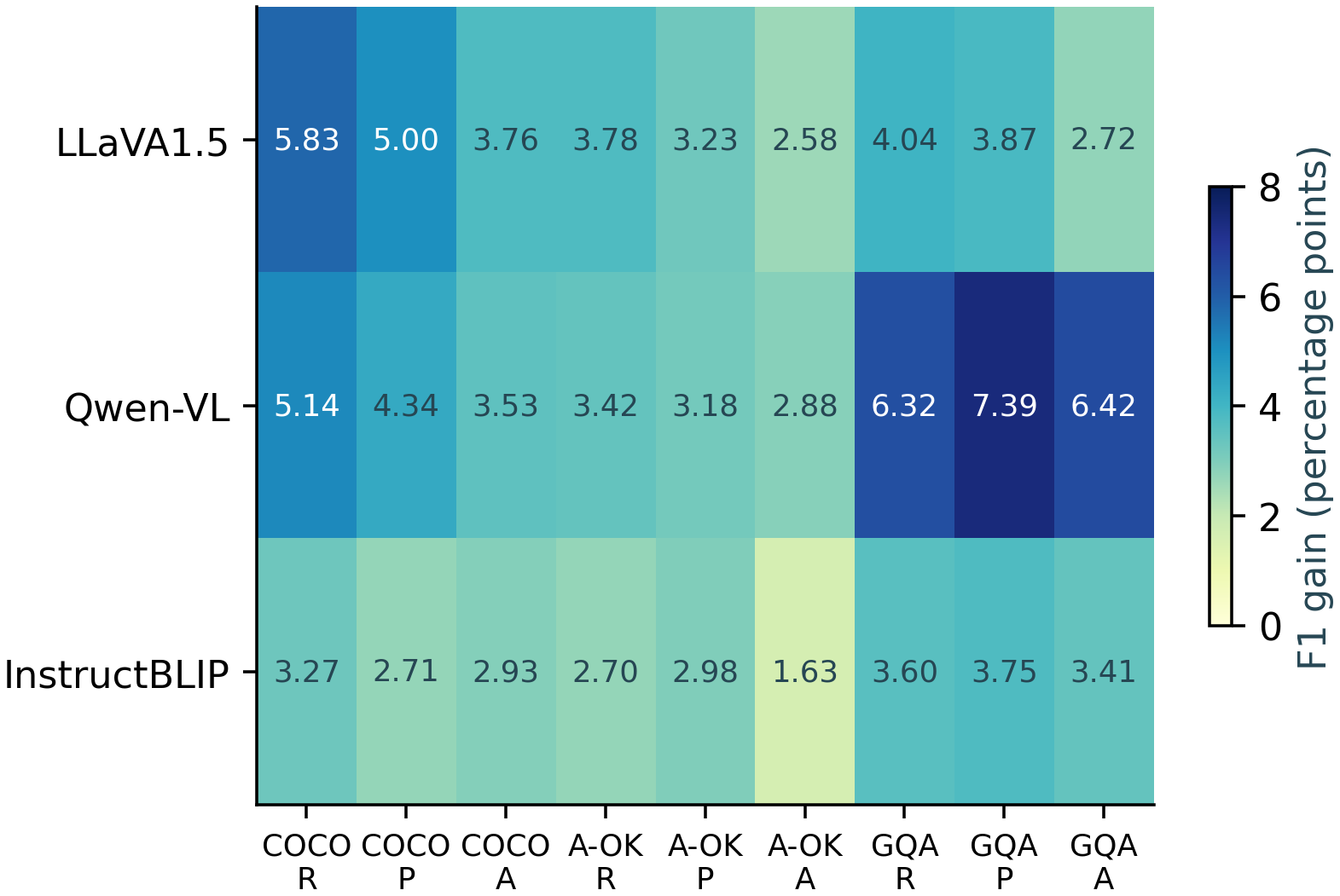}
\caption{F1 gains across sources and samplers.}
\end{subfigure}
\caption{Replots of published VCD Table~1. R, P, and A denote random, popular, and adversarial negatives. Each point or cell derives from the same published backbone pair; none represents EVC+ACC+BSA.}
\label{fig:pope-pair}
\end{figure}

For GQA, Qwen-VL shows a 7.39-point gain under popular negatives and 6.42 under adversarial negatives, while LLaVA1.5's corresponding gains are 3.87 and 2.72. A general statement that more difficult subsets always benefit more would therefore be unsupported. Figure~\ref{fig:pope-pair} visualizes the heterogeneity. The heatmap keeps the dataset and sampling axes separate; pooling them would conceal the interaction we need to understand.

\subsection{MME exposes nonuniform task effects}
Table~\ref{tab:mme} reports the published MME four-task results. LLaVA1.5's total rises from 565.33 to 604.66, Qwen-VL's from 587.33 to 596.67, and InstructBLIP's from 380.33 to 447.67. These gains differ substantially, and the task breakdown reveals an additional limitation: Qwen-VL position declines from 131.67 to 128.00, while InstructBLIP position declines from 66.67 to 64.00. A higher total can therefore coexist with a task-specific regression.

\begin{figure}[!t]
\centering
\begin{subfigure}{0.48\linewidth}\centering
\includegraphics[width=\linewidth]{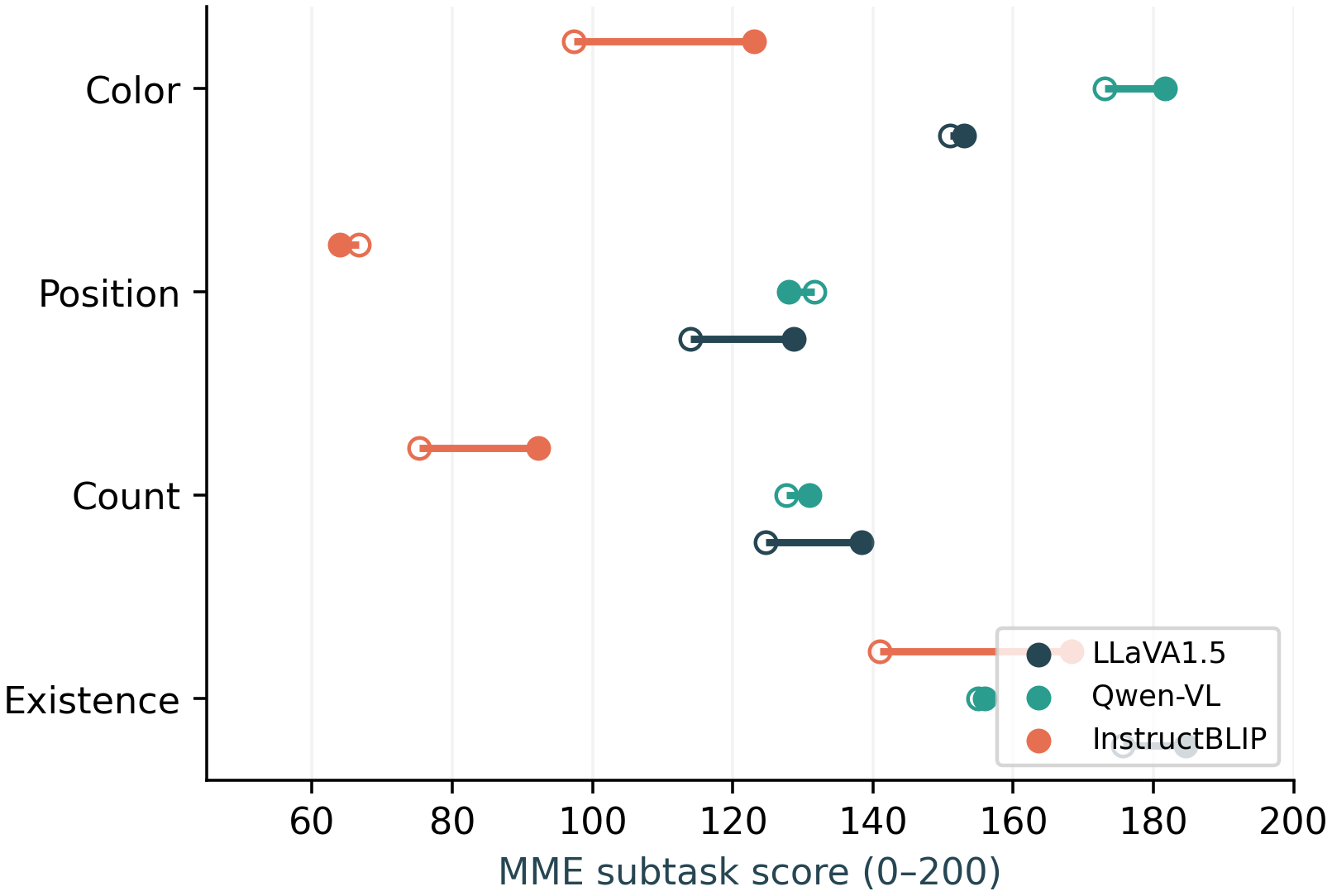}
\caption{MME task scores: open circles regular, filled VCD.}
\end{subfigure}\hfill
\begin{subfigure}{0.48\linewidth}\centering
\includegraphics[width=\linewidth]{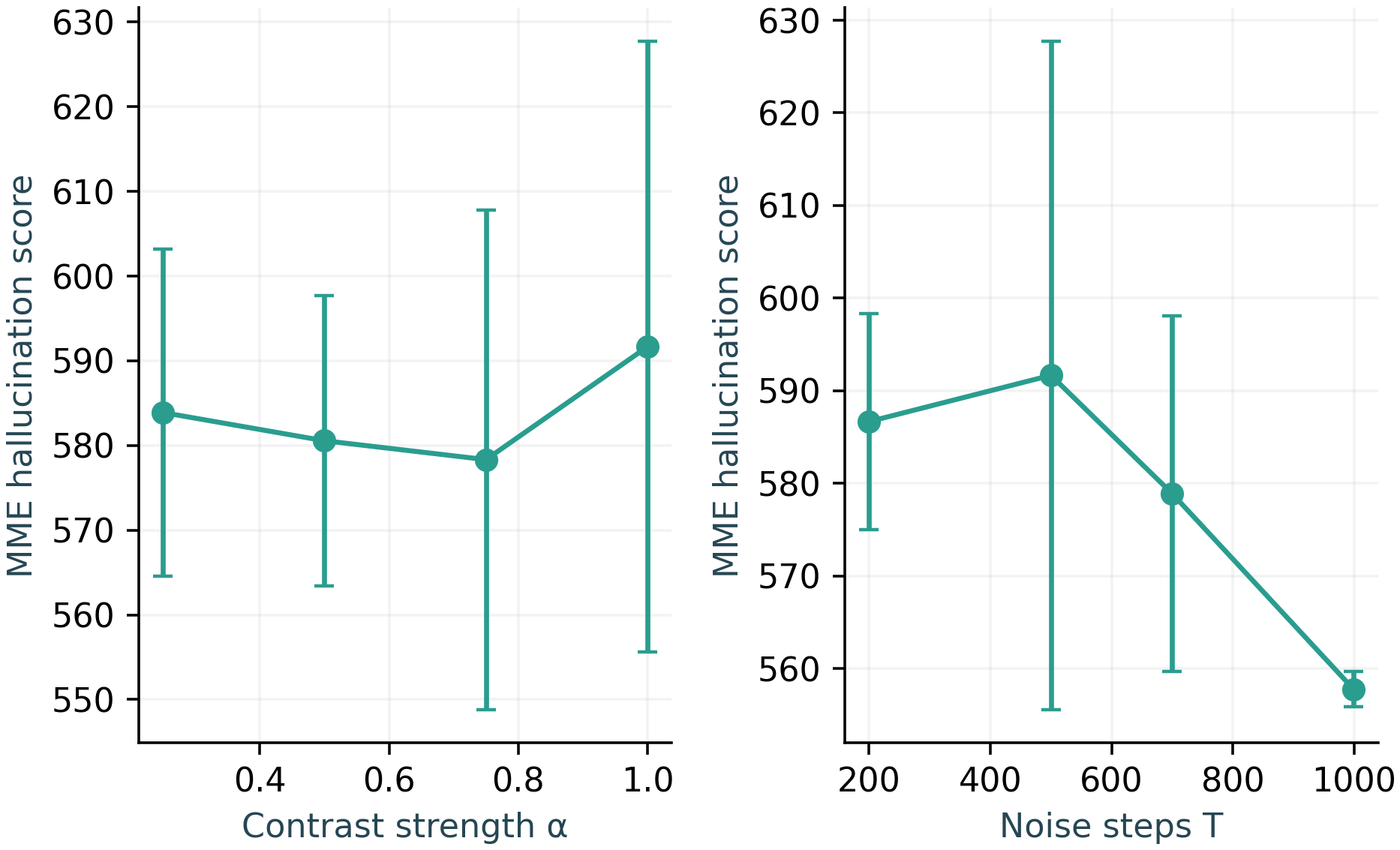}
\caption{Published strength and noise sweeps.}
\end{subfigure}
\caption{Replots of VCD Tables~2, 4, and 5. Error bars on the right reproduce published standard deviations. The left plot exposes task-specific regressions despite improved totals. These are baseline observations, not measurements of the proposed framework.}
\label{fig:mme-pair}
\end{figure}

This matters for local verification. A crop can sharpen evidence that an object exists while deleting the reference needed to judge whether it is on the left. Penalizing disagreement may then either avoid a harmful change or reject a question the global image could answer. The method should be evaluated separately on existence, count, position, and color, with a global-only control for relational tasks. Figure~\ref{fig:mme-pair} shows the published task differences and sensitivity trends that motivate these tests.

\begin{table}[!t]
\centering
\caption{Published MME four-subtask scores, reproduced from VCD Table~2 \citep{vcd2024}. Values are means with original standard deviations; total maximum is 800. No row is a proposed-method measurement.}
\label{tab:mme}
\resizebox{\linewidth}{!}{\begin{tabular}{lllllll}
\toprule
Backbone & Decoding & Existence & Count & Position & Color & Total\\
\midrule
LLaVA1.5 & Regular & $175.67\pm7.51$ & $124.67\pm19.59$ & $114.00\pm9.32$ & $151.00\pm10.45$ & $565.33\pm32.92$\\
LLaVA1.5 & VCD & $184.66\pm6.81$ & $138.33\pm15.68$ & $128.67\pm7.21$ & $153.00\pm7.58$ & $604.66\pm18.76$\\
Qwen-VL & Regular & $155.00\pm3.54$ & $127.67\pm13.36$ & $131.67\pm7.73$ & $173.00\pm9.75$ & $587.33\pm31.06$\\
Qwen-VL & VCD & $156.00\pm6.52$ & $131.00\pm6.19$ & $128.00\pm3.61$ & $181.67\pm5.14$ & $596.67\pm11.61$\\
InstructBLIP & Regular & $141.00\pm13.97$ & $75.33\pm14.16$ & $66.67\pm3.91$ & $97.33\pm16.94$ & $380.33\pm40.20$\\
InstructBLIP & VCD & $168.33\pm11.55$ & $92.33\pm8.47$ & $64.00\pm6.73$ & $123.00\pm11.27$ & $447.67\pm13.36$\\
\bottomrule
\end{tabular}}
\end{table}

The reported totals occasionally differ by 0.01 from sums of rounded subtask means. We preserve the source totals rather than recomputing them from already rounded values. Similarly, the total standard deviation is not the sum or root-sum-square of subtask standard deviations unless their covariance is known. Reconstructing it from the four columns would invent information absent from the publication.

\subsection{Published hyperparameter sensitivity}
Table~\ref{tab:sensitivity} reproduces VCD Tables~4--6 for the noise schedule, contrast strength, and plausibility parameter. These are VCD sensitivity measurements under the source paper's ablation setup, not tests of our gate. For noise steps, the hallucination-subset score is 586.67 at 200, 591.67 at 500, 578.89 at 700, and 557.78 at 999. The perception-subset score instead rises from 1311.47 at 200 to 1345.81 at 999. Optimizing a single aggregate can select a setting that weakens another diagnostic.

The contrast-strength series is not monotonic: hallucination-subset scores are 583.89, 580.56, 578.33, and 591.67 for strengths 0.25, 0.5, 0.75, and 1.0. Their standard deviations overlap substantially. These values motivate sensitivity analysis, but they do not prove a statistically significant optimum. The plausibility sweep likewise changes several scores at once; it cannot be used to infer the causal benefit of adaptive gating without a controlled rerun.

\begin{table}[!t]
\centering
\caption{Published VCD sensitivity data from Tables~4--6 \citep{vcd2024}. H, P, and R denote hallucination, perception, and recognition subsets under the source paper's ablation setup. These scores are not measurements of EVC, ACC, or BSA.}
\label{tab:sensitivity}
\resizebox{0.92\linewidth}{!}{\begin{tabular}{lllll}
\toprule
Parameter & Value & H score & P score & R score\\
\midrule
$T$ & 200 & $586.67\pm11.67$ & $1311.47\pm4.33$ & $338.69\pm19.87$\\
$T$ & 500 & $591.67\pm36.06$ & $1340.89\pm55.91$ & $323.45\pm5.89$\\
$T$ & 700 & $578.89\pm19.17$ & $1339.04\pm40.40$ & $320.95\pm14.18$\\
$T$ & 999 & $557.78\pm1.92$ & $1345.81\pm36.31$ & $321.90\pm10.19$\\
\midrule
$\alpha$ & 0.25 & $583.89\pm19.32$ & $1322.25\pm32.58$ & $330.24\pm13.60$\\
$\alpha$ & 0.5 & $580.56\pm17.11$ & $1315.49\pm27.28$ & $333.45\pm5.77$\\
$\alpha$ & 0.75 & $578.33\pm29.49$ & $1312.93\pm39.31$ & $330.95\pm13.58$\\
$\alpha$ & 1.0 & $591.67\pm36.06$ & $1340.89\pm55.91$ & $323.45\pm5.89$\\
\midrule
$\beta$ & 0 & $577.22\pm11.10$ & $1299.04\pm39.30$ & $302.98\pm19.82$\\
$\beta$ & 0.001 & $574.44\pm6.31$ & $1298.71\pm40.24$ & $289.88\pm14.02$\\
$\beta$ & 0.01 & $583.33\pm18.78$ & $1324.44\pm37.84$ & $327.38\pm17.11$\\
$\beta$ & 0.1 & $591.67\pm36.06$ & $1340.89\pm55.91$ & $323.45\pm5.89$\\
$\beta$ & 0.2 & $591.67\pm7.26$ & $1343.06\pm13.06$ & $328.57\pm16.37$\\
$\beta$ & 0.5 & $635.00\pm7.64$ & $1474.02\pm15.53$ & $331.43\pm13.03$\\
\bottomrule
\end{tabular}}
\end{table}

\section{Evaluation Protocol for the Proposed Framework}
\subsection{Component ablations and falsifiable hypotheses}
Table~\ref{tab:ablation} defines the component study. The EVC-only configuration scores answers without changing tokens, isolating detection from correction. A fixed-strength configuration distinguishes the benefit of local evidence from the benefit of adaptive control. Confidence-based selection is essential because a method can appear safer simply by rejecting low-confidence examples, without using visual verification at all. The full configuration must outperform this control at matched coverage and within the same cost budget to support the proposed mechanism.

\begin{table}[!t]
\centering
\caption{Planned component study. Rows specify interventions, not measured improvements. Every model-performance entry requires a future matched-checkpoint run.}
\label{tab:ablation}
\resizebox{\linewidth}{!}{\begin{tabular}{llll}
\toprule
Configuration & Evidence & Correction & Admission / question tested\\
\midrule
Frozen baseline & None & None & Always answer; original error\\
EVC score only & Multi-view & None & Always answer; error ranking\\
Fixed correction & Multi-view & Fixed strength & Always answer; intervention damage\\
EVC+ACC & Multi-view & Adaptive, bounded & Always answer; correction benefit\\
Confidence selection & None & None & Base confidence; selection control\\
Evidence selection & Multi-view & None & Calibrated score; evidence benefit\\
EVC+ACC+BSA & Multi-view & Adaptive, bounded & Risk and budget; joint behavior\\
\bottomrule
\end{tabular}}
\end{table}

The first hypothesis is that multi-view evidence ranks incorrect answers better than original confidence. It is rejected if its risk--coverage curve fails to improve under a matched budget. The second hypothesis is that adaptive bounded correction reduces harmful changes relative to a fixed-strength local contrast. It is rejected if correct-to-incorrect flips increase without compensating reductions in incorrect-to-correct flips. The third hypothesis is that calibration supports a useful coverage at the target risk. A policy that rejects every example may satisfy a conservative decision rule but supplies no useful answering capability; zero coverage must be treated as an operational failure for that setting.

To examine interactions, record both directions of answer flips, unchanged errors, unchanged correct answers, and abstentions. Do not impose a monotonic ordering on the ablation rows. A module can be redundant, harmful, or beneficial only in combination. The reported full-system score must come from the same run used in the component table; copying a more favorable score from a different seed or parameter selection would invalidate the comparison.

\subsection{Hyperparameters, shift, and efficiency}
Table~\ref{tab:planned} specifies the main sweeps. The view count controls information and cost together. For every view-count condition, report actual auxiliary evaluations, total image encoding work, and peak memory. When comparing with VCD, include a matched-compute version where unused crop budget cannot be silently converted into additional samples from one method alone. The maximum correction strength includes zero with the exact fallback in Eq.~\eqref{eq:fallback}.

\begin{table}[!t]
\centering
\caption{Planned sensitivity, robustness, and cost analysis. Values are evaluation settings; no performance values are implied.}
\label{tab:planned}
\resizebox{\linewidth}{!}{\begin{tabular}{lll}
\toprule
Factor & Prespecified conditions & Required report\\
\midrule
Local view count $K$ & 1, 2, 4, 8 & Risk, coverage, forward equivalents\\
Maximum strength $\alpha_{\max}$ & 0, 0.25, 0.5, 1, 2 & F1, both answer-flip directions\\
Target risk $r_*$ & 0.05, 0.10, 0.15, 0.20 & Coverage, empirical risk, upper bound\\
Image-source shift & COCO calibration to A-OKVQA / GQA & Frozen vs target-calibrated policy\\
Task shift & Existence to count / position / color & Per-task risk and rejection rate\\
Latency and memory & Same hardware, precision, length & p50/p95, first token, peak memory\\
\bottomrule
\end{tabular}}
\end{table}

Calibration can be statistically demanding even before model quality is considered. With $L=20$, $\delta=0.05$, and 500 admitted independent calibration units, the additive term in Eq.~\eqref{eq:ucb} is approximately 0.0774. Thus, an observed risk of 0.02 would have an upper bound near 0.0974; it would not certify a target of 0.05. These values are arithmetic consequences of the stated bound, not measured answer errors. Increasing the number of queries from the same image does not justify treating them as additional independent calibration units.

Efficiency must include preprocessing and evidence acquisition. Record warm-up policy, batch size, synchronization, cache allocation, and answer length. Report both first-token latency and total answer latency, because precomputed crops can move cost before generation. A simple scheduler prioritizes a fixed set of verification positions, while an uncertainty-triggered scheduler is a separate ablation whose calibration policy must be frozen. Compare useful answered questions per unit time in addition to raw tokens per second, since a system that frequently abstains may otherwise appear artificially fast.

Domain-shift testing separates three cases: a policy frozen after source calibration, a policy with target calibration but unchanged scoring parameters, and a fully retuned target policy. The first tests transfer, the second isolates threshold mismatch, and the third estimates adaptation potential at additional data cost. Only the first two preserve the source-trained risk scorer. Their differences should not be merged into a single ``cross-domain'' score.

\subsection{Failure analysis and reporting requirements}
A useful failure audit must include correct predictions damaged by verification. For each audited example, preserve the original answer, corrected answer, released answer or abstention, evidence score, view disagreement, gate strength, and cost. The following categories identify distinct mechanisms rather than interchangeable examples of low accuracy.

\paragraph{Spatial context loss.} A crop containing an object can omit the reference needed for a relation. The global-view control tests whether the local branch introduced the error. Multi-object counting can fail for the same reason when different crops contain overlapping subsets.

\paragraph{Correlated false support.} All views can share a background cue or language prior and produce confident agreement. Low $D_t$ then fails to indicate correctness. Removing the null contrast or perturbing the question wording can help diagnose this behavior, but such probes must not be mistaken for a ground-truth causal explanation.

\paragraph{Null-path artifacts.} A missing visual input can induce an abnormal token distribution. Large evidence lifts can reflect an invalid reference rather than informative pixels. Compare the supported null constructions as an ablation and report failures consistently across backbones.

\paragraph{Affirmation and denial bias.} Suppressing affirmative answers can improve some hallucination statistics while reducing recognition of real objects. Track the Yes ratio, precision, recall, and class-specific rejection rates. Missing evidence in a local crop should increase uncertainty; it should not automatically become a confident negative answer.

\paragraph{Calibration and budget failure.} The admissible set can shrink under domain shift or when the verifier uses most of its budget before reaching a decisive token. Record empty-policy outcomes and budget-triggered abstentions separately. Recalibration can restore a feasible operating point, but it changes the experimental resource assumptions.

These cases should be sampled by a prespecified rule and annotated without selecting only visually persuasive successes. Quantitative category counts require actual model outputs and are therefore not supplied here. The absence of those counts is a limit of the present study, not evidence that a failure mode is rare.

\section{Discussion and Limitations}
The published analysis supports a limited conclusion: a visual contrast can improve several benchmark aggregates while varying in magnitude and producing local regressions. It motivates explicit control over intervention, but it does not show that our particular evidence aggregator, gate, or risk score is optimal. There are plausible alternatives, including global-only verification, attention-space interventions, learned validators, and claim-level admission. The proposed comparison must include strong selection controls before any advantage is attributed to local crops.

The statistical guarantee is deliberately narrower than a deployment assurance. It assumes independent calibration units from the relevant distribution, fixed policies, a fixed correctness rule, and separation from score fitting. Violations can invalidate the certificate even when empirical calibration looks good. The Hoeffding bound is also conservative, particularly at low coverage. More efficient confidence procedures may improve utility, but replacing the bound changes the analysis and requires a fresh check of selection validity.

The token-level formulation introduces an additional modeling gap. Object names can span several tokens, a single token can participate in multiple semantic claims, and answer-level truth does not decompose into independent token decisions. Binary POPE questions provide a tractable initial setting, while open-ended captions need explicit claim extraction and a different correctness audit. The framework should not be advertised as solving arbitrary visual reasoning on the basis of object-presence scores.

All published numbers retain their original provenance, and the four analysis images are deterministic replots of those numbers. The introduction and architecture diagrams illustrate the proposed design. No new model training, inference benchmark, human evaluation, or latency measurement was performed. Consequently, the manuscript makes no state-of-the-art claim, supplies no invented EVC+ACC+BSA score, and leaves its central empirical hypotheses open to the matched evaluation described above.

\section{Conclusion}
We formulated evidence-calibrated selective decoding as a separation of visual support estimation, bounded distribution correction, and budget-aware answer release. EVC measures support across views without treating view agreement as proof, ACC limits changes to plausible candidates and provides an exact fallback, and BSA chooses an admission threshold under explicit calibration assumptions. Re-examining published POPE and MME baselines shows why aggregate gains must be supplemented by per-task analysis, coverage accounting, and intervention-damage checks. The resulting method specification and evaluation protocol are complete enough to implement and falsify. Establishing actual improvements requires future matched-checkpoint experiments; the present evidence establishes the motivation and research design rather than measured superiority.